\documentclass[oneside,onecolumn]{article}

\usepackage[sc]{mathpazo} 
\usepackage[T1]{fontenc} 
\usepackage{microtype} 

\usepackage{parskip} 
\usepackage[english]{babel} 

\usepackage[hmarginratio=1:1,top=32mm,columnsep=20pt,width=18cm]{geometry} 
\usepackage[hang, small,labelfont=bf,up,textfont=it,up]{caption} 
\usepackage{booktabs} 

\usepackage{enumitem} 
\setlist[itemize]{noitemsep} 

\usepackage{abstract} 

\usepackage{titlesec} 
\renewcommand\thesection{\arabic{section}} 
\renewcommand\thesubsection{\arabic{section}.\arabic{subsection}} 
\renewcommand\thesubsubsection{}
\titleformat{\section}[block]{\Large\scshape}{\thesection. }{0em}{} 
\titleformat{\subsection}[block]{\large}{\thesubsection. }{0em}{} 
\titleformat{\subsubsection}[hang]{\normalsize\bfseries}{\thesubsubsection}{0em}{}

\usepackage{fancyhdr} 
\usepackage{titling} 

\usepackage{hyperref} 

\usepackage{numprint} 

\usepackage{listings} 

\usepackage{graphicx} 

\usepackage{mathtools} 

\usepackage{color} 
\usepackage[dvipsnames]{xcolor}

\usepackage{colortbl} 
\definecolor{Gray}{gray}{0.9}
\newcolumntype{g}{>{\columncolor{Gray}}c}

\usepackage{multirow} 

\pretitle{\begin{center}\Huge\bfseries} 
\posttitle{\end{center}} 
\title{Applying Deep Machine Learning for psycho-demographic profiling of Internet users using O.C.E.A.N. model of personality} 
\author{%
\textsc{Iaroslav Omelianenko} \\[1ex] 
\normalsize Research Director, NewGround LLC \\ 
\normalsize \href{mailto:yaric@newground.com.ua}{yaric@newground.com.ua}
}
\date{} 
\renewcommand{\maketitlehookd}{%
\begin{abstract}
\noindent In the modern era, each Internet user leaves enormous amounts of auxiliary digital residuals (footprints) by using a variety of on-line services. All this data is already collected and stored for many years. In recent works, it was demonstrated that it's possible to apply simple machine learning methods to analyze collected digital footprints and to create psycho-demographic profiles of individuals. However, while these works clearly demonstrated the applicability of machine learning methods for such an analysis, created simple prediction models still lacks accuracy necessary to be successfully applied for practical needs. We have assumed that using advanced deep machine learning methods may considerably increase the accuracy of predictions. We started with simple machine learning methods to estimate basic prediction performance and moved further by applying advanced methods based on shallow and deep neural networks. Then we compared prediction power of studied models and made conclusions about its performance. Finally, we made hypotheses how prediction accuracy can be further improved. As result of this work, we provide full source code used in the experiments for all interested researchers and practitioners in corresponding GitHub repository. We believe that applying deep machine learning for psycho-demographic profiling may have an enormous impact on the society (for good or worse) and provides means for Artificial Intelligence (AI) systems to better understand humans by creating their psychological profiles. Thus AI agents may achieve the human-like ability to participate in conversation (communication) flow by anticipating human opponents' reactions, expectations, and behavior.
By providing full source code of our research we hope to intensify further research in the area by the wider circle of scholars.
\end{abstract}
}

\begin{document}

\maketitle


\section{Introduction}

By using various on-line services, modern Internet user leaves an enormous amount of digital tracks in the form of server logs, user-generated content, etc. All these information bits meticulously saved by on-line service providers create the vast amount of digital footprints for almost every Internet user. In recent research \cite{Lambiotte_Kosinski:2014dg}, it was demonstrated that by applying simple machine learning methods it was possible to find statistical correlations between digital footprints and psycho-demographic profile of individuals. The considered psycho-demographic profile comprise of psychometric scores based on five-factor \emph{O.C.E.A.N.} model of personality  \cite{Goldberg:2006dg} and demographic scores such as \emph{Age}, \emph{Gender} and the \emph{Political Views}. The \emph{O.C.E.A.N.} is an abbreviation for \emph{Openness} (Conservative and Traditional - Liberal and Artistic), \emph{Conscientiousness} (Impulsive and Spontaneous - Organized and Hard Working), \emph{Extroversion} (Contemplative - Engaged with outside world), \emph{Agreeableness} (Competitive - Team working and Trusting), and \emph{Neuroticism} (Laid back and Relaxed - Easily Stressed and Emotional).

In this work we decided to test whether applying advanced machine learning methods to analyze digital footprints of Internet users can outperform results of previous research conducted by M. Kosinski: \emph{Mining Big Data to Extract Patterns and Predict Real-Life Outcomes}\cite{Kosinski_Wang_Lakkaraju_Leskovec:2016dg}. For our experiments we used data corpus comprising of  psycho-demographic scores of individuals and their digital footprints in form of Facebook likes. The data corpus used in experiments kindly provided by M. Kosinski through corresponding web site: \url{http://dataminingtutorial.com}.

We started our experiments with building simple machine learning models based on linear/logistic regression methods as proposed by M. Kosinski in \cite{Kosinski_Wang_Lakkaraju_Leskovec:2016dg}. By training and execution of simple models we estimated basic predictive performance of machine learning methods against available data set. Then we continued our experiments with advanced machine learning methods based on shallow and deep neural network architectures.

The full source code of our experiments provided in form of GitHub repository: \url{https://github.com/NewGround-LLC/psistats}

The source code is written in R programming language \cite{R Core Team:2015} which is highly optimized for statistical data processing and allows to apply advanced deep machine learning algorithms by bridging with Google Brain's TensorFlow framework \cite{Google_Brain_Team:2015}.

This paper is organized as follows: In Section~\ref{sec:data_corpus_prepropcessing}, we describe data corpus structure, and necessary data preprocessing steps to be applied. It is followed in Section~\ref{sec:regression_analysis} by details about how to build and run simple prediction models based on linear and logistic regression with results of their execution. In Section~\ref{sec:fcff_ann}, we provide details how to create and execute advanced prediction models based on artificial neural networks. Finally, in Section~\ref{sec:conclusion} we compare the performance of different machine learning methods studied in this work and draw conclusions about the predictive power of studied machine learning models.


\section{Data Corpus Preparation}
\label{sec:data_corpus_prepropcessing}

In this section, we consider the creation of input data corpus from the publicly available data set, and it's preprocessing to allow further analysis by selected machine learning algorithms.


\subsection{Data Set Description }

The data set kindly provided by M. Kosinski and used in this work contains psycho-demographic profiles of \(n_u =\) \numprint{110728} Facebook users and \(n_L =\) \numprint{1580284} of associated Facebook likes. For simplicity and manageability, the sample is limited to U.S. users \cite{Kosinski_Wang_Lakkaraju_Leskovec:2016dg}. The following three files can be downloaded from corresponding web site - \url{http://dataminingtutorial.com}:

\begin{enumerate}
\item \emph{users.csv:} contains psycho-demographic user profiles. It has \(n_u =\) \numprint{110728} rows (excluding the row holding column names) and nine columns: anonymised user ID, gender ("0" for male and "1" for female), age, political views ("0" for Democrat and "1" for Republican), and scores of five-factor model of personality \cite{Goldberg:2006dg}.
\item \emph{likes.csv:} contains anonymized IDs and names of \(n_L =\) \numprint{1580284} Facebook (FB) Likes. It has two columns: ID and name.
\item \emph{users-likes.csv:} contains the associations between users and their FB Likes, stored as user-Like pairs. It has \(nu_L =\) \numprint{10612326} rows and two columns: user ID and Like ID. An existence of a user-Like pair implies that a given user had the corresponding Like on their profile.
\end{enumerate}


\subsection{Data pre-processing}

The raw data preprocessing is an important step in machine learning analysis which will significantly reduce the time needed for analysis and result in better prediction power of created machine learning models.

The detailed description of data corpus preprocessing steps applied during this research given hereafter.


\subsubsection{Construction of sparse users-likes matrix and matrix trimming}
\label{sec:ul_matrix_preprocessing}

To use provided data corpus in machine learning analysis it should be transformed first into optimal format. Taking into account properties of provided data corpus (user can like specific topic only once and most users has generated small amount of likes) its natural to present it as \emph{sparse} matrix  where most of data points is zero (the resulting matrix density is about \numprint{0.006}\% - see Table~\ref{tbl:timmed_matrix}). The \emph{sparse} matrix data structure is optimized to perform numeric operations on sparse data and considerably reduce computational costs compared to the \emph{dense} matrix used for same data set.

After users-likes \emph{sparse} matrix creation, it was \emph{trimmed} by removing rare data points. As a result, the significantly reduced data corpus was created, imposing even lower demands on computational resources and more useful for manual analysis to extract specific patterns. The descriptive statistics of users-likes matrix before and after trimming present in Table~\ref{tbl:timmed_matrix}.

\begin{table}[ht]
\centering
\begin{tabular}{lrr}
\toprule
Descriptive statistics & Raw Matrix & Trimmed Matrix \\
\midrule
\# of users & \numprint{110728} & \numprint{19742} \\ \hline
\# of unique Likes & \numprint{1580284} & \numprint{8523} \\ \hline
\# of User-Like pairs & \numprint{10612326} & \numprint{3817840} \\ \hline
Matrix density & \numprint{0.006}\% & \numprint{2.269}\% \\ \hline
\multicolumn{3}{l}{Likes per User} \\ \hline
\multicolumn{1}{r}{Mean} & 96 & 193 \\ \hline
\multicolumn{1}{r}{Median} & 22 & 106 \\ \hline
\multicolumn{1}{r}{Minimum} & 1 & 50 \\ \hline
\multicolumn{1}{r}{Maximum} & \numprint{7973} & \numprint{2487} \\ \hline
\multicolumn{3}{l}{Users per Like} \\ \hline
\multicolumn{1}{r}{Mean} & 7 & 448 \\ \hline
\multicolumn{1}{r}{Median} & 1 & 290 \\ \hline
\multicolumn{1}{r}{Minimum} & 1 & 150 \\ \hline
\multicolumn{1}{r}{Maximum} & \numprint{19998} & \numprint{8445} \\
\bottomrule
\end{tabular}
\caption{The descriptive statistics of raw and trimmed users-likes matrix with minimum users per like threshold set to \(u_L = 150\) and minimum likes per user \(L_u = 50\)}
\label{tbl:timmed_matrix}
\end{table}

The users-likes matrix can be constructed from provided three comma-separated files with the help of accompanying script written in R language: \emph{src/preprocessing.R}. To use this script make sure that \textbf{input\_data\_dir} variable in the \emph{src/config.R} points to the root directory where sample data corpus in the form of .CSV files were unpacked.

To start preprocessing and trimming, run the following command from terminal in the project's root directory:
\begin{lstlisting}
$Rscript ./src/preprocessing.R \\
         -u 150 -l 50
\end{lstlisting}
where: \textbf{-u} is the minimum number of users per like \(u_L\), and \textbf{-l} is the minimum number of likes per user \(L_u\) to keep in resulting matrix.

The values for the minimum number of users per like \(u_L\) and the minimum number of likes per user \(L_u\) was selected based on recommendations given in \cite{Kosinski_Wang_Lakkaraju_Leskovec:2016dg}. We have experimented with other set of parameters as well (\(u_L = 20\) and \(L_u = 2\)), but accuracy of trained prediction models degraded as result.


\subsubsection{Data imputation of missed values}

The raw data corpus has missed values in column with 'Political' \emph{dependent variable} data. Before building the prediction model for this dependent variable, it is advisable to impute missed values. In this work, we applied multivariate imputation using \textbf{LDA} method with number of multiple imputations equals to \(m = 5\) to fill missed values as described in \cite{Buuren-Groothuis-Oudshoorn:2011dg}.

The data imputation performed by the same \emph{src/preprocessing.R} script, as part of users-likes matrix creation routine. The summary statistics for data imputation applied to \textit{political} variable, presented in Table \ref{tbl:dv_imputation_res}.

\begin{table*}[ht]
\centering
\begin{tabular}{lrrrrrrrrrr}
\toprule
\multicolumn{2}{r}{est} & se & t & df & Pr(>|t|) & lo 95 & hi 95 & nmis & fmi & lambda \\
\midrule
(Intercept) & 1.39 & 0.01 & 102.29 & 1240.53 & 0.00 & 1.36 & 1.41 & NA & 0.06 & 0.05 \\
gender & -0.02 & 0.01 & -2.80 & 25.27 & 0.01 & -0.04 & -0.01 & 0 & 0.44 & 0.40 \\
age & 0.00 & 0.00 & -0.73 & 577.61 & 0.47 & 0.00 & 0.00 & 0 & 0.09 & 0.08 \\
ope & -0.23 & 0.00 & -68.10 & 2446.16 & 0.00 & -0.23 & -0.22 & 0 & 0.04 & 0.04 \\
con & 0.05 & 0.00 & 10.92 & 20.28 & 0.00 & 0.04 & 0.06 & 0 & 0.49 & 0.44 \\
ext & 0.03 & 0.00 & 6.44 & 14.40 & 0.00 & 0.02 & 0.04 & 0 & 0.58 & 0.53 \\
agr & 0.02 & 0.00 & 6.72 & 189.32 & 0.00 & 0.02 & 0.03 & 0 & 0.15 & 0.14 \\
neu & -0.01 & 0.00 & -2.07 & 95.30 & 0.04 & -0.02 & 0.00 & 0 & 0.22 & 0.20 \\
\bottomrule
\end{tabular}
\caption{The descriptive statistics for data imputation applied to \textit{political} variable using \textbf{LDA} method with number of multiple imputations equals to \(m = 5\). The plausibility of applied multivariate imputation can be confirmed by low values in column \textbf{fmi} and \textbf{lambda}. The column \textbf{fmi} contains the \textit{fraction of missing information} as defined in \cite{Rubin:1987dg}, and the column \textbf{lambda} is the proportion of the total variance that is attributable to the missing data \(\lambda = \frac{B + \frac{B}{m}}{T}\)).}
\label{tbl:dv_imputation_res}
\end{table*}


\subsubsection{Dimensionality reduction with SVD}

After two previous steps, the resulting users-likes \emph{sparse} matrix still has a considerable number of features per data sample: \numprint{8523} of feature columns. To make it even more maintainable, we considered applying singular value decomposition \cite{Golub_Reinsch:1970dg}, representing eigendecomposition-based methods, projecting a set of data points into a set of dimensions. As mentioned in \cite{Kosinski_Wang_Lakkaraju_Leskovec:2016dg}, reducing the dimensionality of data corpus has number of advantages:
\begin{itemize}
\item \emph{With reduced features space we can use fewer number of data samples, as it is required by most of the machine learning analysis algorithms that number of data samples exceeds the number of features (input variables)}
\item \emph{It will reduce risk of \textbf{overfitting} and increase statistical power of results}
\item \emph{It will remove \textbf{multicollinearity} and \textbf{redundancy} in data corpus by grouping related features (variables) in single dimension}
\item \emph{It will significantly reduce required computational power and memory requirements}
\item \emph{And finally it makes it easier to analyze data by hand over small set of dimensions as opposite to hundreds or thousands of separate features}
\end{itemize}

To apply \emph{SVD analysis} against generated users-likes matrix run the following command from project's root directory:
\begin{lstlisting}
$Rscript ./src/svd_varimax.R \\
  --svd_dimensions 50 --apply_varimax true
\end{lstlisting}
where: \textbf{--svd\_dimensions} is the number of \emph{SVD dimensions} for projection, and \textbf{--apply\_varimax} is the flag to indicate whether \emph{varimax rotation} should be applied afterwards.


\subsubsection{Factor rotation analysis}

The factor rotation analysis methodology can be used to further simplify \emph{SVD dimensions} and increase their interpretability by mapping the original multidimensional space into a new, rotated space. Rotation approaches can be orthogonal (i.e., producing uncorrelated dimensions) or oblique (i.e., allowing for correlations between rotated dimensions).

In this work during data preprocessing we applied one of the most popular orthogonal rotations - \emph{varimax}. It minimizes both the number of dimensions related to each variable and the number of variables related to each dimension, thus improving the interpretability of the data by human analysts.

For more details on rotation techniques, see \cite{Abdi:2003dg}.


\section{Regression analysis}
\label{sec:regression_analysis}

There is an abundance of methods developed to build prediction machine learning models suitable for analysis of large data sets. They ranging from sophisticated methods such as Deep Machine Learning \cite{Goodfellow-et-al:2016dg}, probabilistic graphical models, or support vector machines \cite{Cortes-Vapnik:1995dg}, to much simpler, such as linear and logistic regressions \cite{Yan-Su:2009dg}. 

Starting with simple methods is a common practice allowing the creation of good baseline prediction model with minimal computational efforts. The results obtained from these models can be used later to debug and estimate the quality of results obtained from advanced models.


\subsection{Regression model description and model-specific data preprocessing}

In our data corpus, we have \emph{eight dependent variables} with psycho-demographic scores of individuals to be predicted. Among those variables, six have continuous values, and two has categorical values. To build the prediction model for variables with continuous values we applied linear regression analysis and for variables with categorical values - logistic regression analysis.

Hereafter we describe the rationale for selection of appropriate regression analysis methods as well as a description of model-specific data preprocessing needed.


\subsubsection{Linear regression analysis}

The linear regression is an approach for modeling the relationship between continuous scalar dependent variable \(y\) and one or more explanatory (or independent) variables denoted \(X\). The case of one explanatory variable is called simple linear regression. For more than one explanatory variable, the process is called multiple linear regression \cite{David-Freedman:2009dg}. In linear regression, the relationships are modeled using linear predictor function \(y = \Theta^TX\) whose unknown model parameters \(\Theta\) estimated from the input data. Such models are called linear models \cite{Hilary-Seal:1967dg}.

We used linear regression to build prediction models for analysis of six continuous dependent variables in given data corpus: \emph{Age}, \emph{Openness}, \emph{Conscientiousness}, \emph{Extroversion}, \emph{Agreeableness}, and \emph{Neuroticism}


\subsubsection{Logistic regression analysis}

The logistic regression is a regression model where the dependent variable is categorical \cite{David-Freedman:2009dg}. It measures the relationship between the categorical dependent variable and one or more independent variables by estimating probabilities using a logistic function \(\sigma(x) = \frac{1}{1 + e^{-x}}\), which is the cumulative logistic distribution \cite{Rodriguez:2007dg}.

We considered only specialized binary logistic regression because categorical dependent variables found in our data corpus (\emph{Gender} and \emph{Political Views}) are binominal, i.e. have only two possible types, "0" and "1".


\subsubsection{Cross-Validation}

We applied \emph{k-fold cross-validation} to help with avoiding model overfitting when evaluating accuracy scores of prediction models. In \emph{k-fold cross-validation}, the original sample is randomly partitioned into k equal sized subsamples. Of the k subsamples, a single subsample is retained as the validation data for testing the model, and the remaining \(k-1\) subsamples are used as training data. The cross-validation process is then repeated \(k\) times (the folds), with each of the \(k\) subsamples used exactly once as the validation data. The k results from the folds can then be averaged to produce a single estimation. The advantage of this method is that all observations are used for both training and validation, and each observation is used for validation exactly once. The 10-fold cross-validation is commonly used, but in general, \(k\) remains an unfixed parameter \cite{Kohavi:1995dg}.


\subsubsection{Dimensionality reduction}

\begin{figure*}
  \includegraphics[width=\textwidth,keepaspectratio=true]{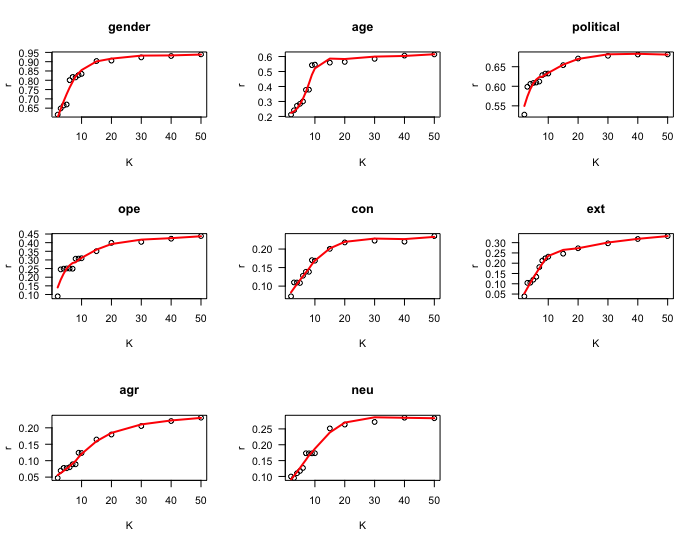}
  \caption{The relationship between the accuracy of predicting psycho-demographic traits and the number of the varimax-rotated singular value decomposition dimensions used for dimensionality reduction. The results suggest that selecting \(K = 50\) SVD dimensions might be a good choice for building models predicting almost all dependent variables, as it offers accuracy that is close to what seems like the higher asymptote for this data. But for \emph{Openness}, \emph{Extroversion}, and \emph{Agreeableness} dependent variables prediction results can be slightly improved with higher numbers of \(K\) SVD dimensions selected.}
  \label{fig:sdv_k_regression_model_correlations}
\end{figure*}

To reduce the number of features (input variables) in the data corpus was applied singular value decomposition (SVD) with subsequent \emph{varimax} factor rotation analysis. The number of the varimax-rotated singular value decomposition dimensions (\(K\)) has a considerable impact on the accuracy of model predictions. To find an optimal number of \emph{SVD dimensions}, we performed analysis of relationships between \(K\) and accuracy of model predictions by creating series of regression models for different values of \(K\). Then we plotted prediction accuracy of regression models against chosen number of \(K\) SVD dimensions. Typically the prediction accuracy grows rapidly within lower ranges of \(K\) and may start decreasing once the number of dimensions becomes large. Selecting value of \(K\) that marks the end of a rapid growth of prediction accuracy values usually offers decent interpretability of the input data topics. In general, the larger \(K\) values often results in better predictive power when preprocessed data corpus further analyzed with specific machine learning algorithm \cite{Zhang-Marron-Shen-Zhu:2007dg}. See Figure~\ref{fig:sdv_k_regression_model_correlations} for results of our experiments.

\begin{figure*}
  \includegraphics[width=\textwidth,keepaspectratio=true]{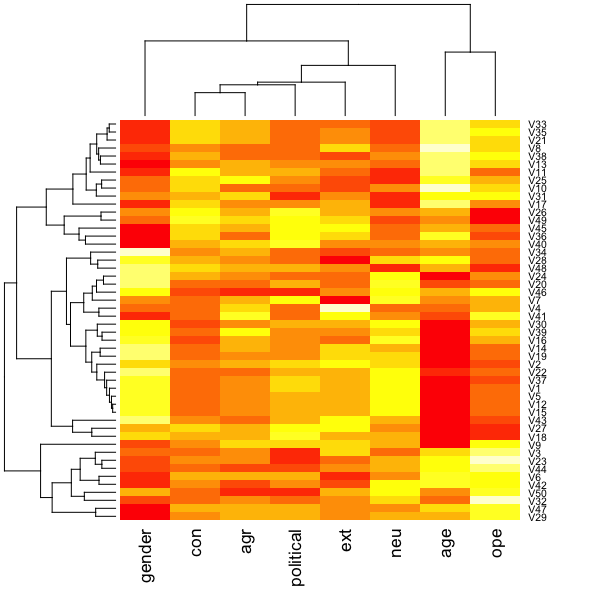}
  \caption{The heat map is presenting correlations between scores of digital footprints of individuals projected on \(K = 50\) varimax-rotated singular value decomposition dimensions and scores of psycho-demographic traits of individuals. The heat map suggests that \emph{Age}, \emph{Gender}, and the \emph{Political view} dependent variables have maximum correlation with a maximal number of SVD dimensions. The higher correlation will result in higher prediction power of regression model for particular dependent variable (which will be shown later).}
  \label{fig:sdv_k_correlation_hmap}
\end{figure*}

To start SVD analysis run the following command from terminal in the project's root directory:
\begin{lstlisting}
$Rscript ./src/analysis.R
\end{lstlisting}
The resulting plots will be saved as "Rplots.pdf" file in the project root and include two plots:
\begin{itemize}
\item the plot with relationships between the accuracy of prediction models for each dependent variable and the number of the varimax-rotated SVD dimensions used for dimensionality reduction (Figure~~\ref{fig:sdv_k_regression_model_correlations}). With this plot, it's easy to visually find an optimal number of \(K\) SVD dimensions to maximize predicting power of regression model per particular dependent variable.
\item the heat map of correlations between scores of digital footprints of individuals projected on specific number of varimax-rotated SVD dimensions and each dependent variable (Figure~\ref{fig:sdv_k_correlation_hmap}). This plot can be used to find most correlated dependent variables visually. Later, it will be shown that predictive models for dependent variables with higher correlation have better prediction accuracy.
\end{itemize}

\subsection{Implementing simple prediction models and its accuracy evaluation}

The given data corpus has eight dependent variables for which to build prediction models. The simple machine learning methods such as regression analysis mostly applied to estimate single dependent variable. But in case when multiple dependent variables need to be estimated the specialized methods of \emph{multivariate} regression analysis can be used. Taking into account that our dependent variables have different types (continuous and nominal) which require different regression analysis methods to be applied, we decided to build separate regression models per each dependent variable. The metric to evaluate accuracy of prediction model is related to the regression method used in the model. In this research we have considered following metrics:
\begin{itemize}
\item the accuracy of prediction model applied to the continuous dependent variable will be measured as Pearson product-moment correlation \cite{Gain:1951dg}
\item the accuracy of prediction model applied to the bi-nominal dependent variable will be measured as area under the receiver-operating characteristic curve coefficient (AUC) \cite{Sing-et-al:2005dg}
\end{itemize}

Before executing models make sure that data corpus already preprocessed as described in Subsection:~"\nameref{sec:ul_matrix_preprocessing}"

When data corpus is ready, the following command can be executed to start linear/logistic regression models building and its predictive performance evaluation (run command from terminal in the project's root directory):
\begin{lstlisting}
$Rscript ./src/regression_analysis.R
\end{lstlisting}
The results will be saved into the file $"out/pred\_accuracy\_regr.txt"$. The prediction accuracy of regression models for data corpus trimmed to contain \(150\) users-per-like and \(50\) likes-per-user and varimax-rotated against \(K = 50\) SVD dimensions presented in Table~\ref{tbl:regression_results}.

\begin{table}[ht]
\centering
\begin{tabular}{llr}
\toprule
Trait & Variable & Pred. accuracy \\
\midrule
Gender & gender & 93.65\% \\
Age & age & 61.17\% \\
Political view & political & 68.36\% \\
Openness & ope & 44.02\% \\
Conscientiousness & con & 25.72\% \\
Extroversion & ext & 30.26\% \\
Agreeableness & agr & 23.97\% \\
Neuroticism & neu & 29.11\% \\
\midrule
\multicolumn{2}{r}{Mean} & 47.03\% \\
\bottomrule
\end{tabular}
\caption{The predictive accuracy of linear and logistic regression models per depended variable (for \(u_L = 150\), \(L_u = 50\), and \(K = 50\) SVD dimensions).}
\label{tbl:regression_results}
\end{table}

From the Table~\ref{tbl:regression_results} it can be seen that prediction accuracy of linear/logistic regression models differs per each dependent variable. Furthermore for most variables the accuracy is too low to be applied in real-life predictions. The most accurate predictions made for \emph{Gender}, \emph{Age}, and \emph{Political view} with \emph{Openness} following after. That correlates well with our previous analysis of SVD correlations heat map (see Figure \ref{fig:sdv_k_correlation_hmap}). In general only prediction model for \emph{Gender} is accurate enough to be useful in real-life applications. Thus, simple linear/logistic regression models can not be used to accurately estimate psycho-demographic profiles of Internet users based only on their Facebook likes.

In following sections, we will test if applying advanced deep machine learning methods can improve prediction accuracy any further.

\section{Fully Connected Feed Forward Artificial Neural Networks}
\label{sec:fcff_ann}

In this work, we considered multilayer \emph{fully connected feed-forward} neural networks (NN) for building simple (shallow) and deep machine learning NN models. The \emph{fully connected} NN characterized by interconnectedness of all units of one layer with all units of the layer before it in the graph. The \emph{feed-forward} NN is not allowed to have cycles from latter layers back to the earlier.

Hereafter we will describe Artificial NN architectures evaluated and prediction accuracy results obtained.

\subsection{The Shallow Feed Forward Artificial NN evaluation}

A \emph{Shallow Neural Network} (SNN) is an artificial neural network with one hidden layer of units (neurons) between input and output layers. Its hidden units (neurons) take inputs from input units (columns of input data matrix) and feeds into the output units, where linear or categorical analysis performed. To mimic biological neuron, hidden units in the neural network apply specific non-linear activation functions. One of the popular activation functions is ReLU non-linearity that we considered as activation function for the units of hidden layers in studied network architectures \cite{Nair-Hinton:2010dg}. It improves information disentangling and linear separability producing efficient variable size representation of model's data. Furthermore ReLU activation is computationally cheaper: there is no need for computing the exponential function as in case of sigmoid activation \cite{Glorot-et-al:2011dg}.

To reduce overfitting was applied dropout regularization with drop-probability \(0.5\), which means that each hidden unit if randomly omitted from the network with the specified probability. This helps to break the rare dependencies that can occur in the training data \cite{Hinton-et-al:2012dg}.

The NN architecture was build using Google Brain's TensorFlow library - an open source software library for numerical computation using data flow graphs. \cite{Google_Brain_Team:2015} Nodes in the graph represent mathematical operations, while the graph edges represent the multidimensional data arrays (tensors) communicated between them. The resulting two layer (one hidden layer) ANN's architecture graph depicted in Figure~\ref{fig:mlp_network_graph}

As the \emph{loss function} to be optimized was selected \emph{Mean Squared Error} (MSE) with \emph{Adam} optimizer (Adaptive Moment Estimation) to estimate it's minimum. The Adam optimizer was selected for \emph{it's proven advantages some of them are that the magnitudes of parameter updates are invariant to rescaling of the gradient, its step sizes are approximately bounded by the step size hyper-parameter, it does not require a stationary objective, it works with sparse gradients, and it naturally performs a form of step size annealing} \cite{Diederik-Ba:2014dg}. The batch size was selected to be \(100\). We also tested training with batch size \(10\) but found no statistically relevant prediction accuracy difference between runs with either batch size but reducing batch size considerably increased training time of NN models.

It was found that optimal number of SVD dimensions for Shallow ANN is \(K = 128\), with number of units in the hidden layer \(512\), and learning rate \(\gamma = 0.0001\). See Table~\ref{tbl:mlp_k_prediction_scores}. 

\begin{table}[ht]
\centering
\begin{tabular}{llcgcc}
\toprule
\multicolumn{2}{l}{} 
& \multicolumn{4}{c}{Prediction accuracy, \(\gamma = 0.0001\)} \\
\cmidrule(r){3-6}
Trait & Variable & \(K = 50\) & \(K = 128\)& \(K = 256\)& \(K = 512\) \\
\midrule
Gender & gender & 93.76\% & 93.61\% & 93.00\% &93.82\%\\
Age & age & 85.00\% & 85.07\% & 85.03\% & 83.59\%\\
Political view & political & 66.65\% & 66.86\% & 67.28\% &65.32\%\\
Openness & ope & 47.22\% & 51.24\% & 47.26\% &44.81\%\\
Conscientiousness & con & 28.56\% & 28.45\% & 29.65\% &28.26\%\\
Extroversion & ext & 32.53\% & 32.87\% & 30.44\% &30.32\%\\
Agreeableness & agr & 25.21\% & 27.66\% & 24.58\% &24.80\%\\
Neuroticism & neu & 33.53\% & 36.24\% & 33.09\% &31.96\%\\
\midrule
\multicolumn{2}{r}{Mean} & 51.58\% & 52.75\% & 51.29\% &50.36\% \\
\bottomrule
\end{tabular}
\caption{The predictive accuracy results of SNN per \(K\) SVD dimensions with learning rate \(\gamma = 0.0001\) and 512 units in the hidden layer.}
\label{tbl:mlp_k_prediction_scores}
\end{table}

The optimal learning rate was selected by comparing ratio of survived hidden units with \emph{non zero ReLU activation} and monitoring loss function over iterations plot (Figure~\ref{fig:mlp_loss_relu-zero_combined}). With presented hyper-parameters it was achieved maximum ratio \(0.57\) of zero ReLU activations for largest learning rate value, which is acceptable taking into account tendency of ReLU to saturate at zero during gradient back propagation stage when strong gradients applied, due to high learning rates \cite{Nair-Hinton:2010dg}. The maximal number of iterations (\numprint{50000}) and correspondingly number of training epochs was selected based on loss function plot (Figure~\ref{fig:mlp_loss_relu-zero_combined}). With series of experiments it was selected as optimal the learning rate  value \(\gamma = 0.0001\), which gives smooth loss function, stable acceptable number of "dead" neurons after ReLU activation (ReLU zero activations ratio is about \(0.5\)), and best prediction scores among runs (see Table~\ref{tbl:mlp_lr_prediction_scores}).

\begin{table}[ht]
\centering
\begin{tabular}{llcgc}
\toprule
\multicolumn{2}{l}{} 
& \multicolumn{3}{c}{Prediction accuracy, \(K = 128\) SVD} \\
\cmidrule(r){3-5}
Trait & Variable & \(\gamma = 0.001\) & \(\gamma = 0.0001\)& \(\gamma = 0.00001\) \\
\midrule
Gender & gender & 93.88\% & 93.61\% & 91.97\%\\
Age & age & 83.68\% & 85.07\% & 73.81\%\\
Political view & political & 66.67\% & 66.86\% & 66.95\%\\
Openness & ope & 49.43\% & 51.24\% & 45.92\%\\
Conscientiousness & con & 27.18\% & 28.45\% & 27.79\%\\
Extroversion & ext & 29.93\% & 32.87\% & 31.93\%\\
Agreeableness & agr & 25.38\% & 27.66\% & 24.51\%\\
Neuroticism & neu & 31.15\% & 36.24\% & 32.24\%\\
\midrule
\multicolumn{2}{r}{Mean} & 50.91\% & 52.75\% & 49.39\% \\
\bottomrule
\end{tabular}
\caption{The predictive accuracy results of SNN per learning rate with \(K = 128\) SVD dimensions and 512 units in the hidden layer.}
\label{tbl:mlp_lr_prediction_scores}
\end{table}

The accompanying launch script provided to conduct experiments under Unix:
\begin{lstlisting}
$./eval_mlp_1.sh ul_svd_matrix_file
\end{lstlisting}
where: \textbf{ul\_svd\_matrix\_file} the path to the preprocessed users-likes matrix with dimension of feature columns reduced as described in \nameref{sec:ul_matrix_preprocessing}

The source code of shallow ANN implementation used for experiment can be found in \emph{src/mlp.R} of accompanying GitHub repository.

\subsection{Feed Forward Deep Learning Networks Architecture Evaluation}

A \emph{Deep Neural Network} (DNN) is an artificial neural network with multiple hidden layers of units between the input and the output layers. The first hidden layer will take inputs from each of the input units and the subsequent hidden layer will take inputs from the outputs of previous hidden layer's units \cite{Bishop:1995dg}. Similar to shallow, deep neural network can model complex non-linear relationships. But added extra layers enable composition of features from lower layers, giving the potential of modeling complex data with fewer units than a similarly performing shallow network \cite{Bengio-Yoshua:2009dg}. \emph{Deep learning discovers intricate structure in large data sets by using the backpropagation algorithm to indicate how a machine should change its internal parameters that are used to compute the representation in each layer from the representation in the previous layer} \cite{LeCun-Bengio-Hinton:2015dg}.

As with shallow ANNs, many issues can arise with training of the deep neural networks, with two most common problems - are overfitting and computation time \cite{Tetko-et-al:1995dg}.

Hereafter we will consider DNN architectures studied in this research.

\subsubsection{The three-layer Deep Learning Network Architecture Evaluation}

Our experiments with deep learning networks we started with simple DNN architecture comprising of two hidden layers with ReLU activation and dropout after each hidden layer with keep probability of \(0.5\). The experimental network graph depicted in Figure~\ref{fig:2dnn_network_graph}.

\begin{table*}[ht]
\centering
\begin{tabular}{llccgcc}
\toprule
\multicolumn{2}{l}{} 
& \multicolumn{4}{c}{Prediction accuracy, \(\gamma = 0.0001\)} \\
\cmidrule(r){3-7}
\multicolumn{2}{l}{} & \(K = 50\) & \(K = 128\)& \(K = 256\)& \(K = 512\)& \(K = 1024\) \\
\cmidrule(r){3-7}
Trait & Variable & 512,256 & 512,256 & 512,256 & 1024,512 & 2048,1024 \\
\midrule
Gender & gender & 92.60\% & 92.14\% & 92.53\% & 93.61\% & 95.68\% \\
Age & age & 85.03\% & 82.97\% & 80.67\% & 81.45\% & 82.53\% \\
Political view & political & 66.66\% & 67.34\% & 67.06\% & 67.26\% & 65.06\% \\
Openness & ope & 44.76\% & 47.56\% & 47.68\% & 46.39\% & 40.61\% \\
Conscientiousness & con & 24.33\% & 25.79\% & 25.68\% & 27.74\% & 24.67\% \\
Extroversion & ext & 30.86\% & 32.35\% & 34.35\% & 28.16\% & 26.53\% \\
Agreeableness & agr & 24.83\% & 25.66\% & 26.48\% & 21.65\% & 29.90\% \\
Neuroticism & neu & 32.17\% & 33.01\% & 35.94\% & 28.50\% & 30.83\% \\
\midrule
\multicolumn{2}{r}{Mean} & 50.15\% & 50.85\% & 51.30\% &49.34\% & 49.48\% \\
\bottomrule
\end{tabular}
\caption{The predictive accuracy results of three layer DNN per \(K\) SVD dimensions with learning rate \(\gamma = 0.0001\), with sizes of hidden layers presented in third table header row as $[hidden_1,hidden_2]$.}
\label{tbl:dnn_k_prediction_scores}
\end{table*}

We started with learning rate \(\gamma = 0.0001\) which resulted in best prediction performance for shallow ANN and attempted series of experiments to estimate optimal value of \(K\) SVD dimensions. The optimal prediction accuracy of DNN model was achieved with \(K = 256\) SVD dimensions and two hidden layers comprising of [\(512,256\)] units correspondingly. Similar prediction accuracy can be achieved with \(K = 1024\) SVD dimensions and [\(2048,1024\)] units per layer with learning rate \(\gamma = 10^{-5}\). But we have not considered later set of hyper-parameters due to its extra computational overhead while giving statistically same results as former set. The results of experiments presented in Table~\ref{tbl:dnn_k_prediction_scores}.

After finding optimal values of \(K\) SVD dimensions and number of units per hidden layer, we have conducted series of experiments to determine optimal \emph{initial} learning rate value for found hyper-parameters. It was experimentally confirmed that \emph{initial} learning rate value \(\gamma = 0.0001\) is the optimal one. See Table~\ref{tbl:dnn_lr_prediction_scores}

\begin{table*}[ht]
\centering
\begin{tabular}{llcgc}
\toprule
\multicolumn{2}{l}{} 
& \multicolumn{3}{c}{Prediction accuracy, \(K = 256\) SVD} \\
\cmidrule(r){3-5}
Trait & Variable & \(\gamma = 0.001\) & \(\gamma = 0.0001\)& \(\gamma = 0.00001\) \\
\midrule
Gender & gender & 90.89\% & 92.53\% & 82.75\%\\
Age & age & 79.60\% & 80.67\% & 79.44\%\\
Political view & political & 67.33\% & 67.06\% & 59.69\%\\
Openness & ope & 44.73\% & 47.68\% & 39.48\%\\
Conscientiousness & con & 21.41\% & 25.68\% & 22.86\%\\
Extroversion & ext & 24.62\% & 34.35\% & 24.07\%\\
Agreeableness & agr & 17.27\% & 26.48\% & 16.34\%\\
Neuroticism & neu & 27.72\% & 35.94\% & 28.53\%\\
\midrule
\multicolumn{2}{r}{Mean} & 46.70\% & 51.30\% & 44.14\% \\
\bottomrule
\end{tabular}
\caption{The predictive accuracy results of three layer DNN per learning rate with \(K = 256\) SVD dimensions, and with sizes of hidden layers 512 and 256 correspondingly.}
\label{tbl:dnn_lr_prediction_scores}
\end{table*}

In our experiments, we applied exponential learning rate decay with the number of steps before decay \numprint{10000} and decay rate of \(0.96\). Such scheme has positive effect on network convergence speed due to the learning rate annealing effect, which gives a system the ability to escape from poor local minima to which it might have been initialized \cite{Kirkpatrick:1983dg}. We selected batch size \(100\) as optimal for this experiment.

The accompanying launch script provided to conduct experiments under Unix:
\begin{lstlisting}
$./eval_dnn.sh ul_svd_matrix_file
\end{lstlisting}
where: \textbf{ul\_svd\_matrix\_file} is the path to the preprocessed users-likes matrix with dimension of feature columns reduced as described in \nameref{sec:ul_matrix_preprocessing}

The source code of DNN with two hidden layers implementation used for experiment can be found in \emph{src/dnn.R} of accompanying GitHub repository.

\subsubsection{The four-layer Deep Learning Network Architecture Evaluation}

This architecture comprise of three hidden layers with ReLU activation and one output linear layer. All network layers are fully connected and network architecture is feed-forward as in all previous NN experiments. The experimental network graph depicted in Figure~\ref{fig:3dnn_network_graph}.

We have tested two dropout regularization schemes: (a) dropout applied after each hidden layer with keep-probability \(0.5\); (b) dropout applied after each second hidden layer with keep probability calculated by formula: \(p_d = \frac{i}{2n}\), (where: \(n\) - number of dropouts, \(i\) - current dropout index). 

It was found that former scheme gives better results than the last one. Thus for final evaluation run, we applied dropout regularization \emph{after each hidden layer}.

\begin{table*}[ht]
\centering
\begin{tabular}{llcccg}
\toprule
\multicolumn{2}{l}{} 
& \multicolumn{3}{c}{Prediction accuracy, \(\gamma = 0.0001\)} \\
\cmidrule(r){3-6}
\multicolumn{2}{l}{} & \(K = 128\) & \(K = 256\)& \(K = 512\)& \(K = 1024\) \\
\cmidrule(r){3-6}
Trait & Variable & 256,128,128 & 512,256,256 & 1024,512,512 & 2048,1024,1024 \\
\midrule
Gender & gender & 55.07\% & 70.63\% & 92.13\% & 91.03\% \\
Age & age & 83.82\% & 82.78\% & 83.80\% & 84.96\% \\
Political view & political & 57.42\% & 61.23\% & 64.52\% & 66.52\% \\
Openness & ope & 13.52\% & 27.50\% & 44.88\% & 44.86\% \\
Conscientiousness & con & 14.41\% & 14.92\% & 20.76\% & 28.72\% \\
Extroversion & ext & 4.06\% & 7.30\% & 22.77\% & 26.85\% \\
Agreeableness & agr & 10.12\% & 10.90\% & 17.69\% & 20.02\% \\
Neuroticism & neu & 8.81\% & 11.98\% & 27.48\% & 28.97\% \\
\midrule
\multicolumn{2}{r}{Mean} & 30.91\% & 35.90\% & 46.75\% & 48.99\% \\
\bottomrule
\end{tabular}
\caption{The predictive accuracy results of three layer DNN per \(K\) SVD dimensions with the learning rate \(\gamma = 0.0001\). The number of units in hidden layers differ per configuration and presented as $[hidden_1,hidden_2]$ in third table header row}
\label{tbl:3dnn_k_prediction_scores}
\end{table*}

Based on our previous experiments with more shallow networks we decided to start with following hyper-parameters: learning rate \(\gamma = 0.0001\), learning rate decay step \numprint{10000} with decay rate \(0.96\), \(K = 128\) SVD dimensions, and hidden layers configuration - [\(256,128,128\)]. 

To find the optimal number of \(K\) SDV dimensions and hidden layers configurations we have conducted series of experiments trying various combinations. The heuristic applied to select the number of units per hidden layer is rather naive and assumes that with dropout probability of \(0.5\) half of the units will be saturated to zero at ReLU activation. Thus we decided to have the number of units in the first hidden layer to be twice as much as the number of features in input data (\(K\)). The results of experiments present in Table~\ref{tbl:3dnn_k_prediction_scores}

Despite the fact that best accuracy was achieved with \(K = 1024\) SDV dimensions and the number of units in hidden layers [\(2048,1024,1024\)] respectively, it was detected slight model overfitting during training/validation with these hyper-parameters applied. So, we have decided to stop increasing \(K\) and number of hidden units as it will give no further gain in prediction accuracy against validation data set and even may lead to worsening of validation accuracy with greater overfitting level. (See Figure~\ref{fig:K_1024_train_loss_overfitting})

The accompanying launch script provided to conduct experiments under Unix:
\begin{lstlisting}
$./eval_3dnn.sh ul_svd_matrix_file
\end{lstlisting}
where: \textbf{ul\_svd\_matrix\_file} is the path to the preprocessed users-likes matrix with dimension of feature columns reduced as described in \nameref{sec:ul_matrix_preprocessing}

The source code of DNN with three hidden layers implementation used for experiments can be found in \emph{src/3dnn.R} of accompanying GitHub repository.

\section{Future Work}

With conducted experiments, we have found that prediction accuracy differs considerably among machine learning methods studied and best results was achieved by using advanced methods based on neural networks architectures. At the same time, the prediction accuracy per individual dependent variable also differs per particular prediction model and selected set of hyper-parameters. From experimental results it can be seen that specific combination of NN architecture with given set of hyper-parameters are best suited for one dependent variable but worsened predictive power for some of the others. 

In future studies, it is interesting to investigate this dependency and build separate NN models per each dependent variable as it was done in case of simple machine learning methods (see Section:~'\nameref{sec:regression_analysis}')

Also, it seems promising to apply methodology described in \cite{Ba-Caruana:2014dg} which provide evidence that shallow networks are capable of learning the same functions as deep learning networks, and often with the same number of parameters as the deep learning networks. In \cite{Ba-Caruana:2014dg} it was shown that with wide shallow networks it's possible to reach the state-of-the-art performance of deep models and reduce training time by the factor of \(10\) using parallel computational resources (GPU).

\section{Conclusion}
\label{sec:conclusion}

From our experiments we found that weak correlation exists between most of \emph{O.C.E.A.N.} psychometric scores of individuals and collected Facebook likes associated with them. Either simple or advanced machine learning algorithms that we have tested provided \emph{poor prediction accuracy} for almost all \emph{O.C.E.A.N.} personality traits. It seems not feasible yet to use machine learning models to accurately estimate \emph{psychometric profile} of an individual based only on Facebook likes. But we believe that by complementing Facebook likes of user with additional data points, it is possible to greatly improve accuracy of machine learning prediction models for psychometric profile estimation.

At the same time, we have found a strong correlation with \emph{demographic traits} of individuals such as \emph{Age}, \emph{Gender}, and the \emph{Political Views} with their Facebook activity (likes). Our experiments confirmed that its possible to use advanced machine learning methods to build the correct demographic profile of an individual based only on collected Facebook likes.

\begin{table*}[ht]
\centering
\begin{tabular}{llcgcc}
\toprule
\multicolumn{4}{l}{} & \multicolumn{2}{c}{DNN} \\
\cmidrule(r){5-6}
\multicolumn{2}{l}{} & Regression & SNN & 512,256 & 2048,1024,1024 \\
\cmidrule(r){3-6}
Trait & Var. & \(K = 50\) & \(K = 128\) & \(K = 256\) & \(K = 1024\) \\
\midrule
Gender & gen. & 93.65\% & 93.61\% & 92.53\% & 91.03\%\\
Age & age & 61.17\% & 85.07\% & 80.67\% & 84.96\%\\
Political view & polit. & 68.36\% & 66.86\% & 67.06\% & 66.52\%\\
Openness & ope & 44.02\% & 51.24\% & 47.68\% & 44.86\%\\
Conscientious. & con & 25.72\% & 28.45\% & 25.68\% & 28.72\%\\
Extroversion & ext & 30.26\% & 32.87\% & 34.35\% & 26.85\%\\
Agreeableness & agr & 23.97\% & 27.66\% & 26.48\% & 20.02\%\\
Neuroticism & neu & 29.11\% & 36.24\% & 35.94\% & 28.97\%\\
\midrule
\multicolumn{2}{r}{Mean} & 47.03\% & 52.75\% & 51.30\% & 48.99\% \\
\bottomrule
\end{tabular}
\caption{The comparison of prediction accuracy for best prediction models found. The best prediction accuracy demonstrated by shallow neural network (SNN) followed by three-layered deep neural network (DNN [512,256]). The further increase of \(K\) SVD dimensions and adding of extra hidden layers lead to model overfitting and degradation of accuracy over validation data set.}
\label{tbl:summary_prediction_results}
\end{table*}

Among all studied machine learning prediction models the best overall accuracy was achieved with Shallow Neural Network architecture. We hypothesize that this may be the result of its ability to learn best parameters space function within an optimal number of \emph{SVD dimensions} applied to the input data set (users-likes sparse matrix). Adding extra hidden layers either leads to model \emph{overfitting} when the number of SVD dimensions is too high, or \emph{underfitting} when the number of SVD dimensions is too low. Also, it's interesting to notice that performance of shallow networks and deep learning networks with two hidden layers are comparable, while with introducing of third and more hidden layers it drops significantly. Thus we can conclude that no further improvements can be gained with extra hidden layers introduced. See Table~\ref{tbl:summary_prediction_results}.

Gathered experimental data confirms that advanced machine learning methods based on variety of studied artificial neural networks architectures outperform simple machine learning methods (linear and logistic regression) described in previous research conducted by M. Kosinski \cite{Kosinski_Wang_Lakkaraju_Leskovec:2016dg}. We believe that further prediction accuracy improvements can be achieved by building separate advanced machine learning models per dependent variable, which is the subject of our future research activities.

\clearpage
\newpage
\appendix
\section{ The SNN evaluation plots}

The following pages provide plots and diagrams related to evaluation of two layer (shallow) feed forward artificial neural network with one fully connected hidden layer and linear output layer.

\begin{figure*}[ht]
  \includegraphics[width=\textwidth,keepaspectratio=true]{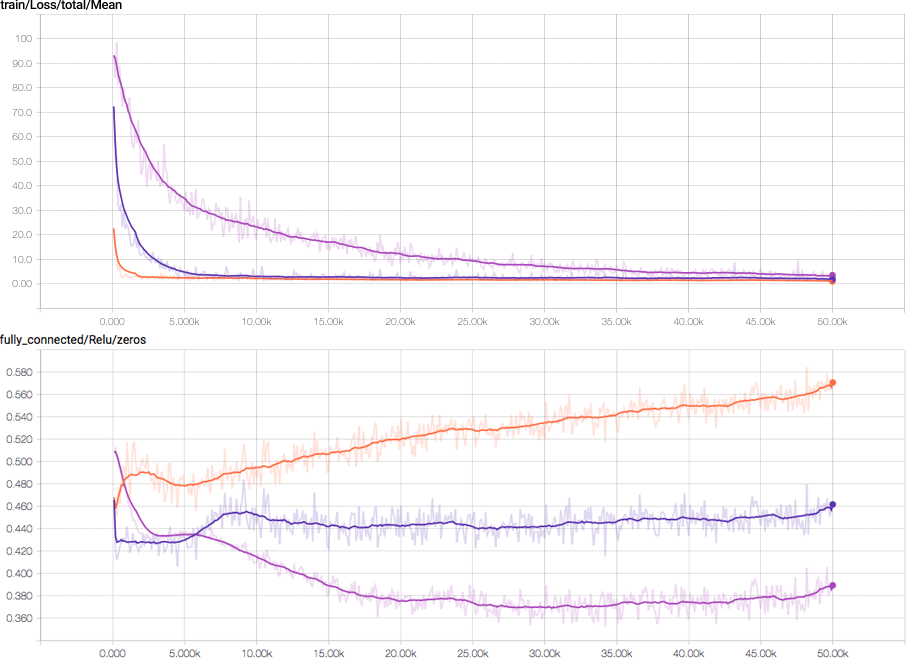}
  \caption{The training process evaluation based on loss values and ReLU zero activations per number of iterations. With higher learning rate (\(\gamma = 0.001\) - \textcolor{BurntOrange}{orange}) we have fast convergence but ratio of ReLU-zero activations is higher than \(0.5\) and quickly rising with relatively low evaluated prediction accuracy, which implies that optimum was missed. With medium learning rate (\(\gamma = 0.0001\) - \textcolor{BlueViolet}{violet}) we have smooth loss function plot with ratio of ReLU-zero activations bellow \(0.5\), giving best prediction scores among all three runs. With lowest learning rate (\(\gamma = 0.00001\) - \textcolor{Purple}{purple}) we can see that learning struggled to find global minimum, reduced speed of convergence, and despite the lowest ReLU zero activations rate - worst prediction accuracy among all runs due to high loss values.}
  \label{fig:mlp_loss_relu-zero_combined}
\end{figure*}

\begin{figure*}[ht]
  \includegraphics[width=\textwidth,keepaspectratio=true]{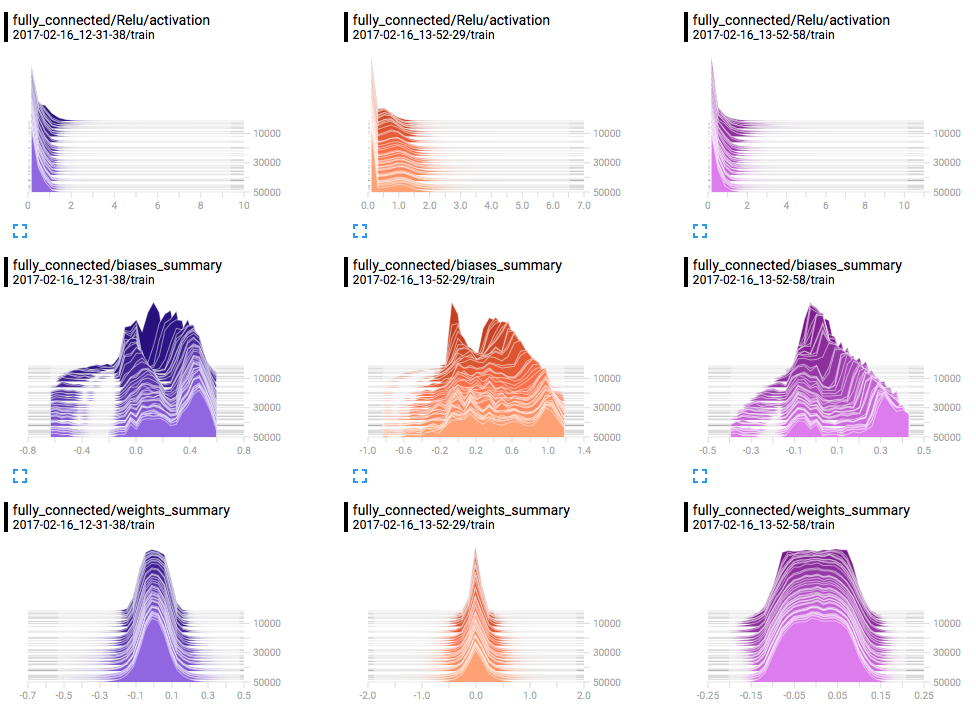}
  \caption{The histograms of various tensors collected during three runs within hidden layer
  (Left: \(\gamma = 0.0001\), middle: \(\gamma = 0.001\), right: \(\gamma = 0.00001\)). By examining weights histograms it may be noticed that middle has widest base with sharp peak which means that layer converged, but search space was widest among all runs. The left one has narrower base and sharp peak, which means that layer converged withing narrower search space and as result has better prediction power. The right one has narrow base but wide plateau at the top, which means that search space is narrow but algorithm still failed to converge. The left and middle histograms has sharp peaks compared to right one, which may be a signal that their learning rate values has more relevance for algorithm convergence and as result we have better predictions for those learning rates.}
  \label{fig:mlp_hidden_hist}
\end{figure*}

\begin{figure*}[ht]
  \includegraphics[width=\textwidth,keepaspectratio=true]{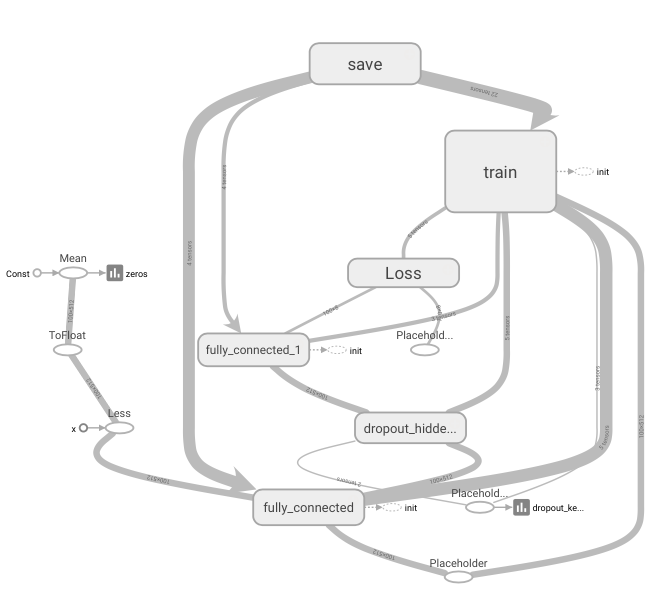}
  \caption{The tensor network graph for multi layer perceptron with one hidden layer (fully\_connected), and one linear output layer (fully\_connected1). The input layer presented as input tensor placeholder. The hidden layer has ReLU activation nonlinearity. The loss function is MSE (mean squared error). The train optimizer is Adam (for Adaptive Moment Estimation).}
  \label{fig:mlp_network_graph}
\end{figure*}
\clearpage
\section{ Deep NN evaluation plots}

The following pages provide plots and diagrams related to evaluation of studied deep neural networks.

\begin{figure*}[ht]
  \includegraphics[width=\textwidth,keepaspectratio=true]{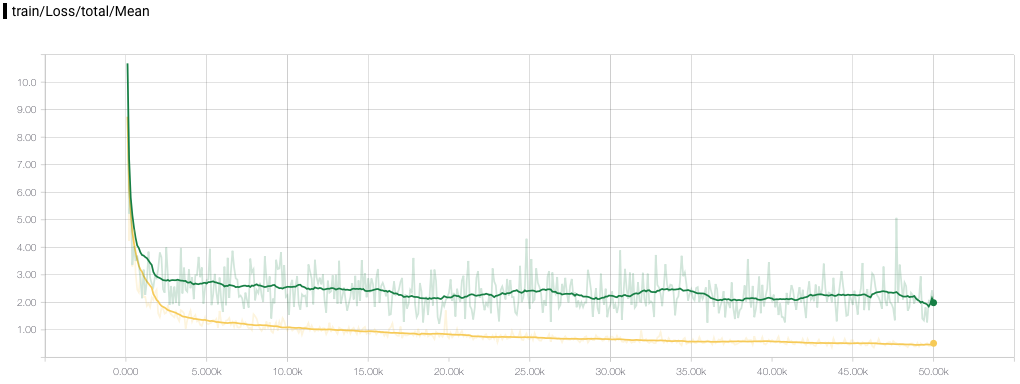}
  \caption{The loss function plot against \textcolor{Dandelion}{train} and \textcolor{OliveGreen}{validation} data sets, for \(K = 1024\) input features and three hidden layers - [2048,1024,1024] units correspondingly. It can be seen that model slightly overfitted against training data - the validation plot is above train plot and doesn't improve with more training steps.}
  \label{fig:K_1024_train_loss_overfitting}
\end{figure*}

\begin{figure*}[ht]
  \includegraphics[width=\textwidth,keepaspectratio=true]{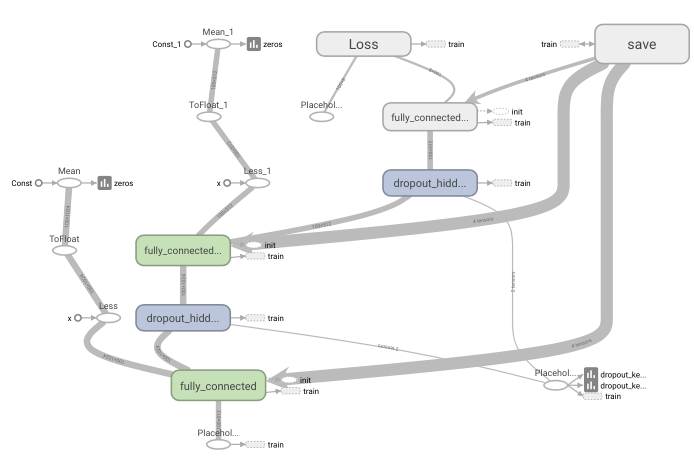}
  \caption{The tensor network graph for DNN with two hidden layers. The input layer presented as input tensor placeholder. The hidden layers has ReLU activation nonlinearity. The loss function is MSE (mean squared error). The train optimizer is Adam (for Adaptive Moment Estimation).}
  \label{fig:2dnn_network_graph}
\end{figure*}

\begin{figure*}[ht]
  \includegraphics[width=\textwidth,keepaspectratio=true]{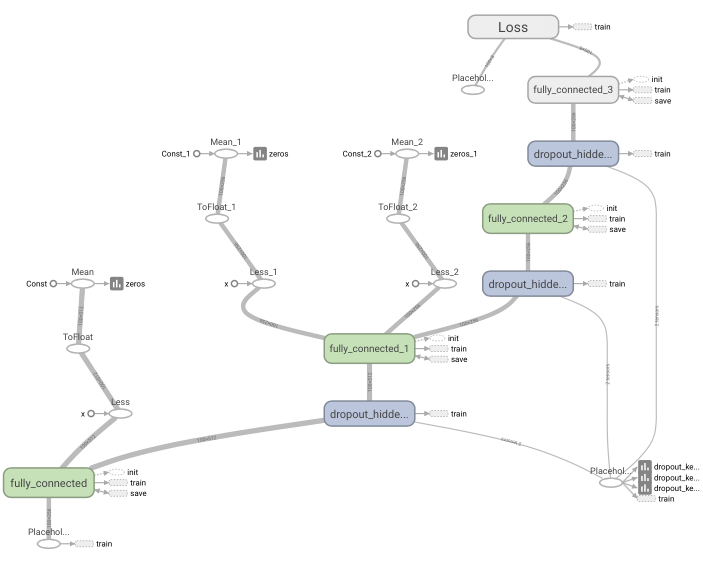}
  \caption{The tensor network graph for DNN with three hidden layers. The input layer presented as input tensor placeholder. The hidden layers has ReLU activation nonlinearity. The loss function is MSE (mean squared error). The train optimizer is Adam (for Adaptive Moment Estimation).}
  \label{fig:3dnn_network_graph}
\end{figure*}



\clearpage
\newpage
\begin{thebibliography}{99} 

\bibitem[Kosinski et. al, 2016]{Kosinski_Wang_Lakkaraju_Leskovec:2016dg}
Michal Kosinski, Yilun Wang, Himabindu Lakkaraju, and Jure Leskovec, (2016).
\newblock Mining Big Data to Extract Patterns and Predict Real-Life Outcomes. 
\newblock {\em Psychological Methods 2016}, Vol. 21, No. 4, 493-506.
\newblock \href{http://dx.doi.org/10.1037/met0000105}{DOI: 10.1037/met0000105}

\bibitem[Lambiotte, R., and Kosinski, M., 2014]{Lambiotte_Kosinski:2014dg}
Lambiotte, R., and Kosinski, M., (2014). 
\newblock Tracking the digital footprints of personality.
\newblock {\em Proceedings of the Institute of Electrical and Electronics Engineers}, 102, 1934-1939. 
\newblock \href{http://dx.doi.org/10.1109/JPROC.2014.2359054}{DOI: 10.1109/JPROC.2014.2359054}

\bibitem[Goldberg et. al, 2006]{Goldberg:2006dg}
Goldberg, L. R., Johnson, J. A., Eber, H. W., Hogan, R., Ashton, M. C., Cloninger, C. R., and Gough, H. G. (2006). \newblock The International Personality Item Pool and the future of public-domain personality measures.
\newblock {\em Journal of Research in Personality}, 40, 84-96.
\newblock \href{http://dx.doi.org/10.1016/j.jrp.2005.08.007}{DOI: 10.1016/j.jrp.2005.08.007}

\bibitem[Golub, G. H., and Reinsch, C. 1970]{Golub_Reinsch:1970dg}
Golub, G. H., and Reinsch, C. (1970).
\newblock Singular value decomposition and least squares solutions.
\newblock {\em Numerische Mathematik}, 14, 403-420.
\newblock \href{http://dx.doi.org/10.1007/BF02163027}{DOI: 10.1007/BF02163027}

\bibitem[Abdi, H., 2003]{Abdi:2003dg}
Abdi, H. (2003).
\newblock Factor rotations in factor analyses.
\newblock In M. Lewis-Beck, A. E. Bryman, \& T. F. Liao (Eds.), {\em The SAGE encyclopedia of social science research methods} (pp. 792-795).
\newblock Thousand Oaks, CA: SAGE.

\bibitem[Goodfellow et al., 2016]{Goodfellow-et-al:2016dg}
Ian Goodfellow and Yoshua Bengio and Aaron Courville, (2016).
\newblock {\em Deep learning.}
\newblock Manuscript in preparation.
\newblock Retrieved from \url{http://www.deeplearningbook.org/}

\bibitem[Daphne Koller, 2012]{Daphne-Koller:2012dg}
\newblock Daphne Koller, (2010-2012)
\newblock {\em Probabilistic Graphical Models.}
\newblock Stanford University.
\newblock Retrieved from \url{http://openclassroom.stanford.edu/MainFolder/CoursePage.php?course=ProbabilisticGraphicalModels}

\bibitem[Cortes \& Vapnik, 1995]{Cortes-Vapnik:1995dg}
\newblock Cortes, C. \& Vapnik, V. (1995).
\newblock Support-vector networks.
\newblock {\em Machine Learning.} 20 (3):273-297. 
\newblock \href{https://dx.doi.org/10.1007/BF00994018}{DOI: 10.1007/BF00994018}

\bibitem[Yan, Su, 2009]{Yan-Su:2009dg}
\newblock Xin Yan, Xiao Gang Su, (2009),
\newblock Linear Regression Analysis: Theory and Computing,
\newblock {\em World Scientific}, pp. 1-2, 
\newblock \href{http://www.worldscientific.com/worldscibooks/10.1142/6986}{ISBN 9789812834119}

\bibitem[David A. Freedman, 2009]{David-Freedman:2009dg}
\newblock David A. Freedman (2009).
\newblock Statistical Models: Theory and Practice.
\newblock {\em Cambridge University Press}, p. 26.

\bibitem[Hilary L. Seal, 1967]{Hilary-Seal:1967dg}
\newblock Hilary L. Seal (1967).
\newblock The historical development of the Gauss linear model.
\newblock {\em Biometrika.} 54 (1/2): 1-24. 
\newblock \href{https://dx.doi.org/10.1093/biomet/54.1-2.1}{DOI: 10.1093/biomet/54.1-2.1}

\bibitem[Rodriguez, G., 2007]{Rodriguez:2007dg}
\newblock Rodriguez, G. (2007).
\newblock {\em Lecture Notes on Generalized Linear Models.} pp. Chapter 3, page 45.
\newblock Retrieved from \url{http://data.princeton.edu/wws509/notes/}

\bibitem[Sing, Sander, Beerenwinkel, Lengauer, 2005]{Sing-et-al:2005dg}
\newblock Sing, T., Sander, O., Beerenwinkel, N., \& Lengauer, T. (2005).
\newblock ROCR: Visualizing classifier performance in R.
\newblock {\em Bioinformatics}, 21, 3940-3941.
\newblock \href{http://dx.doi.org/10.1093/bioinformatics/bti623}{DOI: 10.1093/bioinformatics/bti623}

\bibitem[Gain, 1951]{Gain:1951dg}
\newblock Gain, A. K. (1951).
\newblock The frequency distribution of the product moment correlation coefficient in random samples of any size draw from non-normal universes.
\newblock {\em Biometrika.} 38: 219-247. 
\newblock \href{https://dx.doi.org/10.1093/biomet/38.1-2.219}{DOI: 10.1093/biomet/38.1-2.219}

\bibitem[Kohavi, Ron, 1995]{Kohavi:1995dg}
\newblock Kohavi, Ron (1995).
\newblock A study of cross-validation and bootstrap for accuracy estimation and model selection.
\newblock {\em Proceedings of the Fourteenth International Joint Conference on Artificial Intelligence.}
\newblock San Mateo, CA: Morgan Kaufmann. 2 (12): 1137-1143.
\newblock \href{https://citeseerx.ist.psu.edu/viewdoc/summary?doi=10.1.1.48.529}{CiteSeerX 10.1.1.48.529}

\bibitem[Zhang, Marron, Shen,\& Zhu, 2007]{Zhang-Marron-Shen-Zhu:2007dg}
\newblock Zhang, L., Marron, J., Shen, H., \& Zhu, Z. (2007).
\newblock Singular value decomposition and its visualization.
\newblock {\em Journal of Computational and Graphical Statistics}, 16, 833-854.

\bibitem[van Buuren, Groothuis-Oudshoorn, 2011]{Buuren-Groothuis-Oudshoorn:2011dg}
\newblock Stef van Buuren and Karin Groothuis-Oudshoorn (2011).
\newblock mice: Multivariate Imputation by Chained Equations in R.
\newblock {\em Journal of Statistical Software} 45 (3)
\newblock American Statistical Association.
\newblock Retrieved from \url{http://doc.utwente.nl/78938/}

\bibitem[Rubin DB, 1987]{Rubin:1987dg}
\newblock Rubin DB (1987).
\newblock {\em Multiple Imputation for Nonresponse in Surveys.}
\newblock John Wiley \& Sons, New York.

\bibitem[Christopher M. Bishop,	1995]{Bishop:1995dg}
\newblock Christopher M. Bishop (1995).
\newblock {\em Neural Networks for Pattern Recognition},
\newblock Oxford University Press, Inc. New York, NY, USA \copyright1995 
\newblock ISBN:0198538642

\bibitem[Bengio, Yoshua, 2009]{Bengio-Yoshua:2009dg}
\newblock Bengio, Yoshua (2009).
\newblock Learning Deep Architectures for AI.
\newblock {\em Foundations and Trends in Machine Learning.} 2 (1): 1-127. 
\newblock \href{https://dx.doi.org/10.1561/2200000006}{DOI: 10.1561/2200000006}

\bibitem[LeCun, Bengio, Hinton, 2015]{LeCun-Bengio-Hinton:2015dg}
\newblock LeCun, Yann; Bengio, Yoshua; Hinton, Geoffrey (2015).
\newblock Deep learning.
\newblock {\em Nature.} 521: 436–444.
\newblock \href{https://dx.doi.org/nature14539}{DOI:10.1038/nature14539} 

\bibitem[Tetko, Livingstone, Luik, 1995]{Tetko-et-al:1995dg}
\newblock Tetko, I. V.; Livingstone, D. J.; Luik, A. I. (1995). 
\newblock Neural network studies. 1. Comparison of Overfitting and Overtraining. 
\newblock {\em J. Chem. Inf. Comput. Sci.} 35 (5): 826-833.
\newblock \href{https://dx.doi.org/10.1021/ci00027a006}{DOI: 10.1021/ci00027a006}

\bibitem[Hinton, G. et al., 2012]{Hinton-et-al:2012dg}
\newblock Hinton, Geoffrey E.; Srivastava, Nitish; Krizhevsky, Alex; Sutskever, Ilya; Salakhutdinov, Ruslan R. (2012).
\newblock {\em Improving neural networks by preventing co-adaptation of feature detectors.}
\newblock arXiv preprint \href{https://arxiv.org/abs/1207.0580}{arXiv:1207.0580v1}

\bibitem[Nair, Hinton, 2010]{Nair-Hinton:2010dg}
\newblock Vinod Nair and Geoffrey Hinton (2010).
\newblock Rectified linear units improve restricted Boltzmann machines. 
\newblock {\em ICML.}
\newblock Retrieved from \url{http://machinelearning.wustl.edu/mlpapers/paper_files/icml2010_NairH10.pdf}

\bibitem[Glorot, Bordes, Bengio, 2011]{Glorot-et-al:2011dg}
\newblock Xavier Glorot, Antoine Bordes, Yoshua Bengio (2011).
\newblock Deep Sparse Rectifier Neural Networks.
\newblock {\em JMLR W\&CP} 15:315-323
\newblock Retrieved from \url{http://jmlr.org/proceedings/papers/v15/glorot11a.html}

\bibitem[Kingma, Ba, 2014]{Diederik-Ba:2014dg}
\newblock Diederik P. Kingma; Lei Jimmy Ba (2014). 
\newblock {\em Adam: A method for stochastic optimization.}
\newblock arXiv preprint \href{https://arxiv.org/abs/1412.6980}{arXiv:1412.6980}

\bibitem[Ba, Caruana, 2014]{Ba-Caruana:2014dg}
\newblock Lei Jimmy Ba, Rich Caruana (2014).
\newblock {\em Do Deep Nets Really Need to be Deep?}
\newblock arXiv preprint \href{https://arxiv.org/abs/1312.6184}{arXiv:1312.6184}

\bibitem[Kirkpatrick et al., 1983]{Kirkpatrick:1983dg}
\newblock S. Kirkpatrick, C. D. Gelatt Jr., M. P. Vecchi (1983).
\newblock Optimization by Simulated Annealing.
\newblock {\em Science}, 13 May 1983: Vol. 220, Issue 4598, pp. 671-680
\newblock \href{https://dx.doi.org/10.1126/science.220.4598.671}{DOI: 10.1126/science.220.4598.671}

\bibitem[R Core Team, 2015]{R Core Team:2015}
R Core Team, (2015).
\newblock R: A language and environment for statistical computing. 
\newblock {\em Vienna, Austria: R Foundation for Statistical Computing.} 
\newblock Retrieved from \url{http://www.R-project.org/}

\bibitem[Google Brain Team, 2015]{Google_Brain_Team:2015}
Google Brain Team, (2015).
\newblock TensorFlow is an open source software library for numerical computation using data flow graphs. 
\newblock Retrieved from \url{https://www.tensorflow.org}
 
\end{thebibliography}
\end{document}